\documentclass[10pt,twocolumn,letterpaper]{article}

\usepackage{cvpr}              

\usepackage{graphicx}
\usepackage{amsmath}
\usepackage{amssymb}
\usepackage{booktabs}
\usepackage[pagebackref,breaklinks,colorlinks]{hyperref}
\usepackage{enumitem, array}
\usepackage[dvipsnames]{xcolor}
\usepackage[capitalize]{cleveref}
\crefname{section}{Sec.}{Secs.}
\Crefname{section}{Section}{Sections}
\Crefname{table}{Table}{Tables}
\crefname{table}{Tab.}{Tabs.}

\def\cvprPaperID{5762} 
\def\confName{CVPR}
\def\confYear{2022}

\begin{document}

\title{Unsupervised Homography Estimation with Coplanarity-Aware GAN}

\author{
    Mingbo Hong$^{1,2}$\thanks {Equal contribution. $^{\dagger}$Corresponding authors.}\ \ \ \ Yuhang Lu$^{1,3*}$\ \ \ \ Nianjin Ye$^{1}$\ \ \ \ Chunyu Lin$^{4}$\ \ \ \ Qijun Zhao$^{2\dagger}$\ \ \ \ Shuaicheng Liu$^{5,1\dagger}$
  \\
    $^{1}$ Megvii Technology \quad  $^{2}$ Sichuan University  \quad  $^{3}$ Univesity of South Carolina  \\
    $^{4}$ Beijing Jiaotong University \quad  $^{5}$ University of Electronic Science and Technology of China \\
}

\maketitle

\begin{abstract}
Estimating homography from an image pair is a fundamental problem in image alignment. Unsupervised learning methods have received increasing attention in this field due to their promising performance and label-free training. However, existing methods do not explicitly consider the problem of plane-induced parallax, which will make the predicted homography compromised on multiple planes. In this work, we propose a novel method HomoGAN to guide unsupervised homography estimation to focus on the dominant plane. First, a multi-scale transformer network is designed to predict homography from the feature pyramids of input images in a coarse-to-fine fashion. Moreover, we propose an unsupervised GAN to impose coplanarity constraint on the predicted homography, which is realized by using a generator to predict a mask of aligned regions, and then a discriminator to check if two masked feature maps are induced by a single homography. To validate the effectiveness of HomoGAN and its components, we conduct extensive experiments on a large-scale dataset, and results show that our matching error is 22\% lower than the previous SOTA method. Code is available at \url{https://github.com/megvii-research/HomoGAN}.
\end{abstract}
\vspace{-4mm}

\section{Introduction}
\label{sec:intro}
Homography estimation is a fundamental computer vision problem that plays an important role in a wide range of applications, such as image/video stitching~\cite{ZaragozaAPAP13, guo2016joint}, camera calibration~\cite{zhang2000flexible}, HDR imaging~\cite{gelfand2010multi} and SLAM~\cite{Mur-ArtalMT15, zou2012coslam}.
It is defined as the estimation of the projective transformation between two views on the same plane in 3D space~\cite{shao2021localtrans}.
Traditional methods typically address this problem by following a pipeline of feature extraction~\cite{LoweSIFT04, BayETGSURF08, RubleeRKB11ORB}, correspondence matching, and solving direct linear transform~\cite{daglib_AHAZ} with outlier rejection~\cite{FischlerRansac81}. 
But these methods often suffer from the lack of discriminative keypoints when dealing with textureless or blurry images.

\begin{figure}[t]
\centering
   \includegraphics[width=1.0\linewidth]{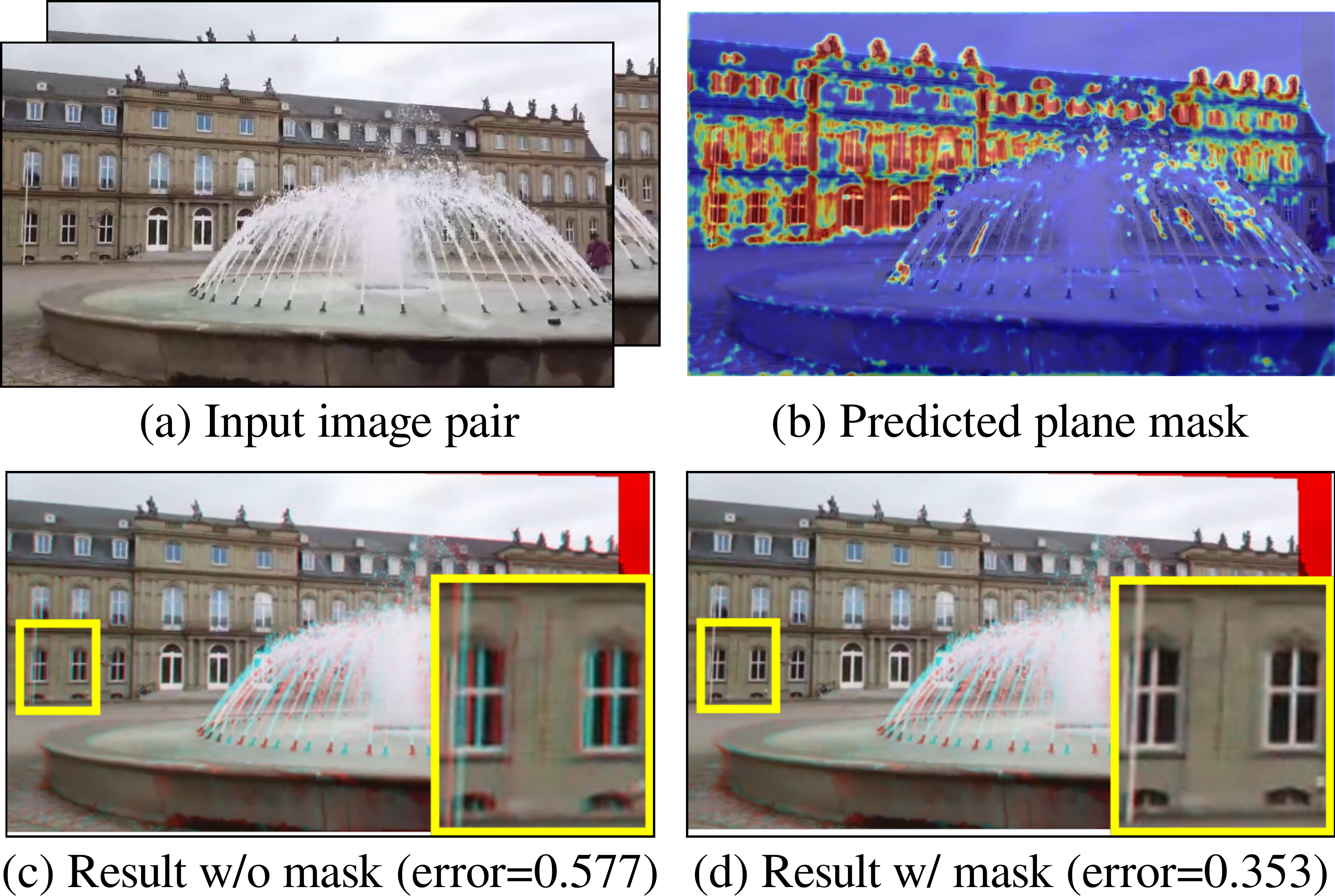}
   \vspace{-6mm}
   \caption{Predicting homography for images with a multi-plane scene will lead to virtual parallax. We propose an unsupervised homography estimation method that enforces the model to focus on the dominant plane by leveraging coplanarity constraint, thus significantly reducing the matching error. (c) and (d) are generated by superimposing the warped source image on the target image.}
\label{fig:teaser}
\vspace{-5mm}
\end{figure}

Recently, unsupervised learning methods have gained popularity in homography estimation~\cite{detone2016deep, NguyenCSTK18, zhang2020content, ye2021motion, shao2021localtrans}.
These methods directly predict the homography from a pair of source and target images using a neural network, of which an important optimization objective is to minimize the distance from the warped source image to the target image.
They do not rely on keypoints, and could perform better than traditional methods in textureless scenarios.
However, when there exist multiple planes in the scene, optimizing over the entire image will lead to a compromised result, \textit{i.e.}, the predicted homography is averaged on all planes and not accurate on the dominant plane, as shown in Fig.~\ref{fig:teaser}.
Note that the planes of interest are not limited to rigid planes such as grounds, buildings, and walls, but also include planes that can be approximately induced by a homography, such as mountains in the distance.
Some existing methods propose to remove large foregrounds or moving objects from the input images by predicting a mask~\cite{zhang2020content, LeLZA20}.
But their masks are implicitly optimized as a side product of homography estimation and lack explicit guidance, thus cannot address the \textit{plane-induced parallax}.

In this work, we introduce an unsupervised approach to empower homography estimators to focus on a dominant plane instead of the entire image.
Assume a scene with multiple planes, we cannot align the entire image with one homography unless two images are related by a conjugate rotation~\cite{daglib_AHAZ}.
Instead, we can obtain a mask to indicate if each pixel is well aligned by the predicted homography.
If the homography is induced by the dominant plane, then the aligned regions should be on the same plane and occupy a significantly large area.
With this knowledge, our main idea is to impose additional \textit{coplanarity constraint} and area penalty on the aligned regions of the mask.

To achieve this goal, we propose a new method \textit{HomoGAN} with two unique designs.
First, to guide the model to focus on the dominant plane, we introduce an unsupervised GAN to impose coplanarity constraint, in which the generator predicts soft masks of aligned regions from a pair of feature maps, while the discriminator checks if the masked features is coplanar.
Together with the foreground area constraint, the generated masks are expected to highlight the dominant plane, which can in turn guide the training of homography estimator.
Second, a multi-scale transformer network is designed to predict the homography from a pair of feature pyramids in a coarse-to-fine fashion.
Compared to CNN-based alternatives, the query-key correlation of transformers is more natural to establish local correspondence for homography estimation.
In sum, this work makes the following contributions:
\begin{itemize}[topsep=0mm]
\itemsep -1.1mm
    \item We propose a coplanarity-aware GAN to address the problem of plane-induced parallax for homography estimation without ground truth.
    \item We design a coarse-to-fine homography estimation transformer with self-attention encoders for capturing local correspondences and class-attention decoders for summarizing global information.
    \item Our method achieves the state-of-the-art performance on unsupervised homography estimation, and outperforms previous methods by $22\%$ on matching error.
\end{itemize}

\section{Related Work}
\label{sec:related}

\paragraph{Traditional Homography Estimation}
A traditional pipeline of homography estimation usually involves steps of feature extraction~\cite{LoweSIFT04, BayETGSURF08, RubleeRKB11ORB}, feature matching, and solving direct linear transform~\cite{daglib_AHAZ} with outlier rejection~\cite{FischlerRansac81}. 
Classic feature extraction methods include SIFT~\cite{LoweSIFT04}, SURF~\cite{BayTGSURF06, BayETGSURF08}, ORB~\cite{RubleeRKB11ORB}, LPM~\cite{ma2019localityLPM}, GMS~\cite{bian2017gms}, BEBLID~\cite{suarez2020beblid} etc. 
Recently, a number of learning-based features are proposed, such as LIFT~\cite{YiTLF16}, SuperPoint~\cite{detone2018superpoint}, SOSNet~\cite{TianSOSNet19} and OAN~\cite{zhang2019learningOAN}. 
There are also deep learning approaches for feature matching, including SuperGlue~\cite{sarlin2020superglue}, LoFTR~\cite{sun2021loftr}, etc. 
Finally, outliers should be rejected for robust estimation, where RANSAC~\cite{FischlerRansac81}, MAGSAC~\cite{BarathMAGSAC19} and IRLS~\cite{1977RobustIRLS} are widely used.

\vspace{-4mm}
\paragraph{Deep Homography Estimation}
Deep homography estimation can be categorized into supervised and unsupervised methods.
Supervised methods~\cite{detone2016deep, LeLZA20, shao2021localtrans} learn from image pairs with ground truth homographies, which are difficult to obtain for natural images in the wild.
If learning from synthetic images, the lack of realistic transformation will degrade their generalization ability.
Unsupervised methods~\cite{NguyenCSTK18, zhang2020content, ye2021motion} typically optimize their model by minimizing a distance from the source image warped by the predicted homography to the target image.
\cite{zhang2020content} and \cite{LeLZA20} introduced mask prediction into homography estimation, but their goal is to remove large foregrounds or moving objects, while our goal is to preserve a single dominant plane with explicit constraint.
Recently, Shao \textit{et al.}~\cite{shao2021localtrans} proposed a supervised transformer for cross-resolution homography estimation.
However, aiming at different tasks, our architecture designs are also different, where they propose a transformer with local attention, while ours contains a self-attention encoder and class-attention decoder.

\vspace{-4mm}
\paragraph{Dominant Plane Detection}
Detecting dominant planes in images has been studied in past literature.
For example, Conrad \textit{et al.}~\cite{conrad2010homography} proposed a homography-based method to detect the ground plane for robot navigation.
In~\cite{de2016dominant}, a learning-based method was proposed to recognize dominant planes in indoor scenes.
More recently, \cite{yang2018recovering, liu2018planenet, liu2019planercnn, tan2021planetr} proposed to detect and recover 3D planes from a single image using various neural networks.
However, these methods are not applicable to our problem because they either work only on rigid physical planes or require a large number of ground truth masks for training.
In contrast, our plane detection GAN could help homography estimators to concentrate on the dominant plane without direct supervision.

\section{Method}
\label{sec:method}

\begin{figure*}
	\centering
	\includegraphics[width=\linewidth]{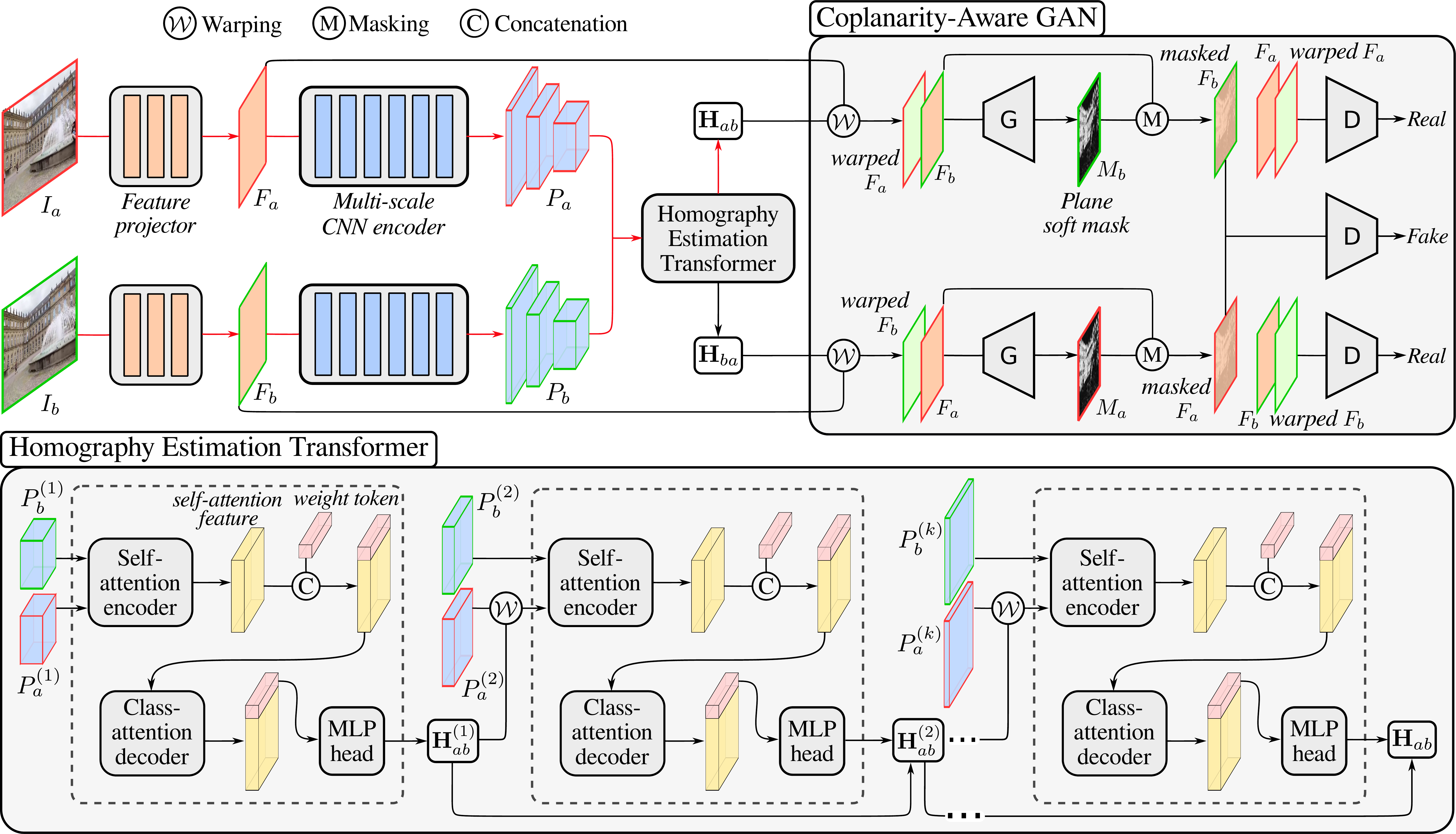}
    \vspace{-6mm}
	\caption{The overall pipeline of HomoGAN. Our network architecture consists of four modules: 1) Feature projector. A CNN module that projects the input images to a shallow feature space. 2) Multi-scale CNN encoder. A CNN module that generates a feature pyramid for each image. 3) Homography Estimation Transformer. A transformer with cascaded encoder-decoder blocks that predicts the homography from coarse to fine. 4) Coplanarity-Aware GAN. An adversarial network that imposes coplanarity constraint on the model by predicting soft masks of the dominant plane. Red arrows indicate the inference pipeline.}
\label{fig:pipeline}
\vspace{-4mm}
\end{figure*}

\subsection{Overview}
In this section, we introduce a new method HomoGAN for unsupervised homography estimation with small baseline. 
Given a pair of gray-scale image patches $I_a$ and $I_b$ of size $H \times W$, we predict the homography transformation from $I_a$ to $I_b$, denoted as $\mathbf{H}_{ab}$.
Following~\cite{ye2021motion}, we decompose a homography matrix into 8 orthogonal flow bases, and predict the weights of $8$ bases instead of regressing the homography matrix or the corner offsets~\cite{detone2016deep,NguyenCSTK18, zhang2020content}.
The pipeline of our method is illustrated in Fig.~\ref{fig:pipeline}.

We first employ a \textbf{feature projector} $\mathcal{F}(\cdot)$ to convert the input images $I_a$ and $I_b$ to feature maps $F_a$ and $F_b$, where $\mathcal{F}$ is a lightweight CNN module with three basic convolutional blocks.
This module does not change the input dimension, \textit{i.e.}, $F_* \in \mathbb{R}^{1 \times H \times W}$.
The intention is to project images to a shallow feature space that is robust to luminance variations~\cite{zhang2020content}, such that the following steps could focus on geometric transformations. 
Subsequently, a \textbf{multi-scale CNN encoder} is employed to prepare feature pyramids for coarse-to-fine homography estimation. 
The encoder is composed of $k$ cascaded convolutional blocks, where each block downsamples the input by a scale of $2$ and outputs one level of the feature pyramid.
We denote the feature pyramids as $P_a$ and $P_b$, and the $i$-th level of $P_*$ as $P_*^{(i)}$, of which the feature size is $\frac{H}{2^{k-i+1}} \times \frac{W}{2^{k-i+1}}$.

At the core of our method are the newly-proposed \textbf{Homography Estimation Transformer} and \textbf{Coplanarity-Aware GAN}.
The former is a transformer network that is specifically designed for homography estimation, which consumes the extracted feature pyramids $P_a$ and $P_b$, and predicts the homography from coarse to fine. 
The latter is an plug-in module that can be applied to any homography estimation networks to impose coplanarity constraint. 
It could guide the model to focus on the dominant plane in $I_a$ and $I_b$ by predicting soft plane masks via unsupervised adversarial learning.
Finally, the entire model is optimized by minimizing a hybrid unsupervised objective function.


\subsection{Homography Estimation Transformer}
Given a pair of feature pyramids $P_a$ and $P_b$, we propose a transformer network to estimate the underlying homography transformation. 
The design of the transformer adopts a coarse-to-fine strategy. 
We start from the top-level feature $P_{*}^{(1)}$, and progressively estimate the homography at a finer scale until $P_{*}^{(k)}$, as shown in Fig.~\ref{fig:pipeline}.

The homography refinement is realized by $k$ cascaded transformer modules with independent weights, denoted as $\mathcal{T}_1, \mathcal{T}_2, \cdots , \mathcal{T}_k$, respectively.
In the $i$-th transformer module, we first warp the feature map $P_{a}^{(i)}$ using the previous $\mathbf{H}_{ab}^{(i-1)}$, and then $\mathcal{T}_i$ takes $P_{b}^{(i)}$ and the warped $P_{a}^{(i)}$ as inputs, and predicts their homography transformation.
Finally, $\mathbf{H}_{ab}$ is updated by accumulating the output of $\mathcal{T}_i$ to the previous result with a weight of the current scale.
This process can be formulated as:
\begin{equation}
\label{eq:H}
\mathbf{H}_{ab}^{(i)} = \mathbf{H}_{ab}^{(i-1)} + 2^{k-i+1} \cdot \mathcal{T}_i(\mathcal{W}(\mathbf{H}_{ab}^{(i-1)}, P_{a}^{(i)}), P_{b}^{(i)}),
\end{equation}
where $i \in [1,k]$, $\mathcal{W}$ is the warping operation, and $\mathbf{H}_{ab}^{(0)}$ is an identical transformation. 
We add ${H}_{ab}$ together as they are in the form of flow bases.
Similarly, we can compute the homography $H_{ba}$ from $I_b$ to $I_a$ by swapping $P_a$ and $P_b$.

Within each transformer module, we use an encoder-decoder architecture to compute the homography at a specific scale level, which consists of a self-attention encoder, a class-attention decoder, and an MLP head.

\vspace{-4mm}
\paragraph{Self-attention encoder}
The role of the encoder is to encode the feature correspondences into an intermediate embedding. 
We first concatenate the warped $P_{a}^{(i)}$ and $P_{b}^{(i)}$ to obtain a new feature of size $\mathbb{R}^{2C_i\times(H_i W_i)}$.
Swin Transformer~\cite{liu2021swin} is employed as the backbone of our encoder because it computes window-based self-attention instead of global attention, which is computationally efficient and suitable for capturing local feature correspondences.
As opposed to its original design, we use it for global-to-local refinement instead of local-to-global abstraction.
In the $i$-th module, we use $(i-1)$ patch merging layers to downsample the input feature to keep the dimension of output self-attention feature to be $2C_1 \times (H_1 W_1)$ for all modules, which will facilitate the following decoding.
We also replace the pixel shuffle operation in patch merging with a convolutional block to enhance local information exchange.



\vspace{-4mm}
\paragraph{Class-attention decoder}
In the decoding stage, we summarize task-specified information from the general-purpose self-attention features. 
Inspired by the intermediate class token~\cite{touvron2021cait}, we introduce a weight token into the model, which is a learnable tensor of size $2C_i \times 8$.
It is concatenated with the self-attention feature to construct a feature of size $2C_1 \times (H_1 W_1+8)$, and then fed into class-attention sub-blocks to compute the attention between the weight token and the self-attention feature, such that the weight token collects information of all patches to predict homography flow weights.
Since all self-attention feature have the same size, we maintain a single weight token throughout the network, but do not need to reinitialize it in every module.


Finally, we fetch the processed weight token and use an MLP head with two linear layers to project it to a weight vector of length $8$, which is the result of of the $i$-th module $\mathcal{T}_i(\cdot)$, and it is used to update the homography in Eq.~(\ref{eq:H}).

\subsection{Coplanarity-Aware GAN}
If without any constraint, the aforementioned transformer will consider all regions in $I_a$ and $I_b$ when computing $\mathbf{H}_{ab}$ and $\mathbf{H}_{ba}$, which might not be desired for homography estimation when there exist multiple planes. 
To let the transformer focus on the dominant plane, we propose an unsupervised GAN to detect the dominant plane by leveraging coplanarity constraint, as illustrated in Fig.~\ref{fig:pipeline}.

First, we apply the predicted $\mathbf{H}_{ab}$ and $\mathbf{H}_{ba}$ to $F_a$ and $F_b$, respectively, and obtain the warped feature maps $F_a^{'}$ and $F_b^{'}$.
By contrasting $F_a$ against $F_b^{'}$ or $F_b$ against $F_a^{'}$, we can check if a region is well aligned by the predicted homographies.
In the ideal case, the aligned regions are located in the dominant plane of the scene.
To realize this, we employ a generator network $\mathcal{G}$ to check the spatial consistency between a pair of feature maps.
It generates a soft mask that highlights regions that are well aligned.
The architecture of $\mathcal{G}$ is composed of three convolutional layers with an ASPP~\cite{chen2017deeplab} module inserted.
We obtain two masks $M_a$ and $M_b$ using $\mathcal{G}$, where $M_a = \mathcal{G}(F_a, F_b^{'})$ and $M_b = \mathcal{G}(F_b, F_a^{'})$.

For planes in general position, the homography is determined
uniquely by the plane and vice versa~\cite{daglib_AHAZ}.
Therefore, if the foreground of $M_a$ and $M_b$ are in the dominant plane, the induced homography is unique.
Motivated by this, we design a discriminator network $\mathcal{D}$ which is expected to discriminate if the transformation within the input pair is a single homography.
We take $(F_a, F_a^{'})$ and $(F_b, F_b^{'})$ as real pairs, of which the unique homographies are $\mathbf{H}_{ab}$ and $\mathbf{H}_{ba}$, respectively.
The masked $F_a$ and $F_b$, \textit{i.e.}, $(M_a F_a, M_b F_b)$ is taken as the fake pair.
The discriminator $\mathcal{D}$ is constructed by $7$ convolutional layers and a global average pooling layer.
Through adversarial training, we implicitly impose the coplanarity constraint on $M_a$ and $M_b$. 

Following Wasserstein GAN-GP~\cite{gulrajani2017WGANGP}, we utilize the Wasserstein distance to measure the discrepancy between real and fake pairs to stabilize the training. 
We also adopt the gradient reversal layer~\cite{ganin2015GRL} for one-stage adversarial training, and the adversarial loss is written as:
\begin{equation}
\label{eq:adv}
    L_{adv} = \mathcal{D}(M_a F_a, M_b F_b) - (\mathcal{D}(F_a, F_a^{'}) + \mathcal{D}(F_b, F_b^{'})),
\end{equation}
where the sign of the gradient of $\mathcal{D}(M_a F_a, M_b F_b)$ is reversed in backpropagation.

To stabilize training, a gradient penalty term \cite{gulrajani2017WGANGP} is applied to $\mathcal{D}$ to enforce the Lipschitz constraint:
\begin{equation}
    L_{gp} = \mathcal{E}((\left \| \bigtriangledown_\mathcal{D} \right \|_{2}-1)^{2}),
\end{equation}
where $\mathcal{E}$ is the mean function and $\bigtriangledown$ is the gradient operator.

Furthermore, we compute the cross-entropy loss between $M_a$, $M_b$ and a constant mask $\hat{M}$ as an auxiliary loss:
\begin{equation}
L_{aux} = \textup{CE}(M_a, \hat{M}) + \textup{CE}(M_b, \hat{M}),
\end{equation}
which encourages $M_a$ and $M_b$ to have larger foregrounds while keeping the coplanarity.

Finally, the loss function of the plane detection GAN is:
\begin{equation}
    L_{plane} = \alpha_{1}L_{adv} + \alpha_{2}L_{gp} + \alpha_{3}L_{aux},
\end{equation}
where $\alpha_{1}$, $\alpha_{2}$ and $\alpha_{3}$ are the weights of each term set as 0.01, 10 and 0.1, respectively.

\subsection{Network Training}
Besides the plane detection loss $L_{plane}$, we also minimize two other unsupervised losses for network training.
The first one is an alignment loss $L_{align}$ to compare feature maps before and after warping by the predicted homographies. 
We first compute the pixel-wise triplet loss~\cite{zhang2020content} to obtain a distance map $G_{ab}$ by:
\begin{equation}
\label{eq:triplet}
     G_{ab} =\max(|| F_{a}^{'} - F_b ||_1 - || F_a-F_b ||_1 + 1, 0),
\end{equation}
Similarly, a distance map $G_{ba}$ can be obtained by replacing $F_{a}^{'}-F_b$ with $F_{b}^{'}-F_a$.
We further apply the predicted masks to $G_{ab}$ and $G_{ba}$ to emphasize the dominant plane, and compute $L_{align}$ by:
\begin{equation}
\label{eq:align}
    L_{align} = \frac{\sum _i M_a^{'} M_b G_{ab}}{\sum _i M_a^{'} M_b} + \frac{\sum _i M_a M_b^{'} G_{ba}}{\sum _i M_a^{'}M_b},
\end{equation}
where $M_*^{'}$ are the warped masks and $i$ is the pixel index.

The second term is a feature identity loss $L_{FIL}$ that lets the feature projector $\mathcal{F}$ be warp-equivalent~\cite{ye2021motion}, which is written as:
\begin{equation}
\begin{aligned}
    L_{FIL} = 
    &\left \| \mathcal{W}(\mathbf{H}_{ab}, \mathcal{F}(I_a))-\mathcal{F}(\mathcal{W}(\mathbf{H}_{ab}, I_a)) \right \|_1 + \\ 
    &\left \| \mathcal{W}(\mathbf{H}_{ba}, \mathcal{F}(I_b))-\mathcal{F}(\mathcal{W}(\mathbf{H}_{ba}, I_b)) \right \|_1
\end{aligned}
\end{equation}
It forces $\mathcal{F}$ to filter luminance variations while keeping the geometric transformation.

Finally, the overall loss function is  written as:
\begin{equation}
\label{eq:total}
L_{total}= L_{align} + L_{FIL} + L_{plane}.
\end{equation}

To achieve the best performance, we adopt a two-stage strategy for network training.
We first leave out the GAN part and only train the remaining parts because we empirically find that abnormal homography predictions in the early stage may lead to unstable adversarial training.
In this stage, the masks in Eq. (\ref{eq:align}) and $L_{plane}$ in Eq. (\ref{eq:total}) are temporarily disabled.
When the first stage converges, we add the coplanarity-aware GAN back to the model, and enable all loss terms to start the second stage training.

\begin{figure}
	\centering
	\includegraphics[width=\linewidth]{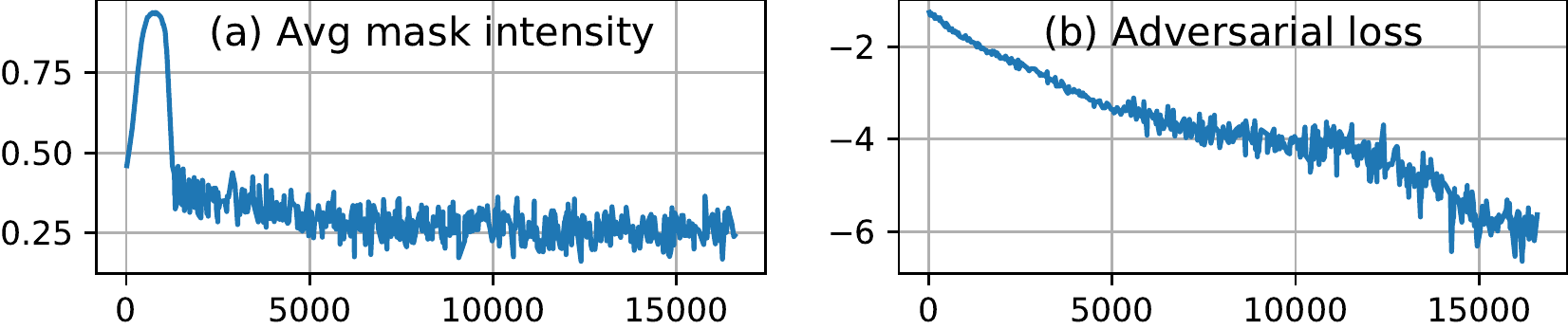}
	\vspace{-6mm}
	\caption{Average mask intensity and adversarial loss in training.}
\label{fig:mask}
\vspace{-4mm}
\end{figure}

\vspace{1mm}
\noindent\textbf{Discussion}
When constructing real pairs in the GAN, we do not apply the predicted mask to them to avoid degenerate solutions, \textit{i.e.}, $\mathcal{G}$ simply generates all-zero masks.
One may question that the appearance discrepancy between real and fake pairs will distract $\mathcal{D}$ from discriminating the coplanarity.
However, the GAN is not trained standalone, but as a regularization of the transformer.
If $\mathcal{D}$ simply discriminates by appearance discrepancy, $\mathcal{G}$ will output all-one masks, then it has zero impact on the main objective $L_{align}$.
To reach the global optimum, the optimizer will guide $\mathcal{D}$ to discriminate geometric discrepancy, so that $\mathcal{G}$ can output masks of coplanar regions.
To justify it, we visualize the average mask intensity and the adversarial loss in training in Fig.~\ref{fig:mask}.
We can see that the mask intensity first boosts to $\sim$1, and then falls to 0.2-0.4, which means that $\mathcal{G}$ is first biased to all-one masks, but then corrected to output plane masks.
Meanwhile, the adversarial loss keeps declining, indicating the effectiveness of our training strategy.

\section{Experiments}
\label{sec:exp}

\begin{figure*}[t]
	\centering
	\includegraphics[width=\linewidth]{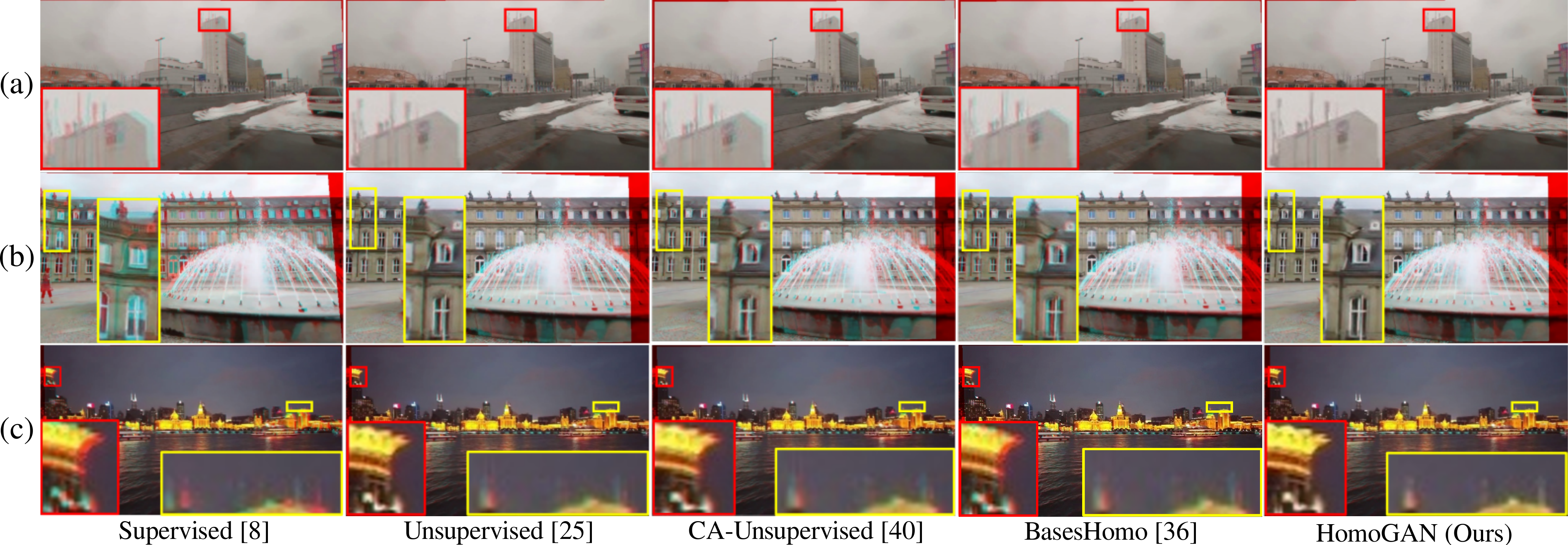}
    \vspace{-6mm}
	\caption{Qualitative results of our method and four other deep learning-based methods. Images are generated by superimposing the warped source images on the target image. Error-prone regions are highlighted with red and yellow boxes. Best viewed with zooming in.}
\label{fig:DNNbased}
\vspace{-2mm}
\end{figure*}


\begin{figure}[t]
	\centering
	\includegraphics[width=\linewidth]{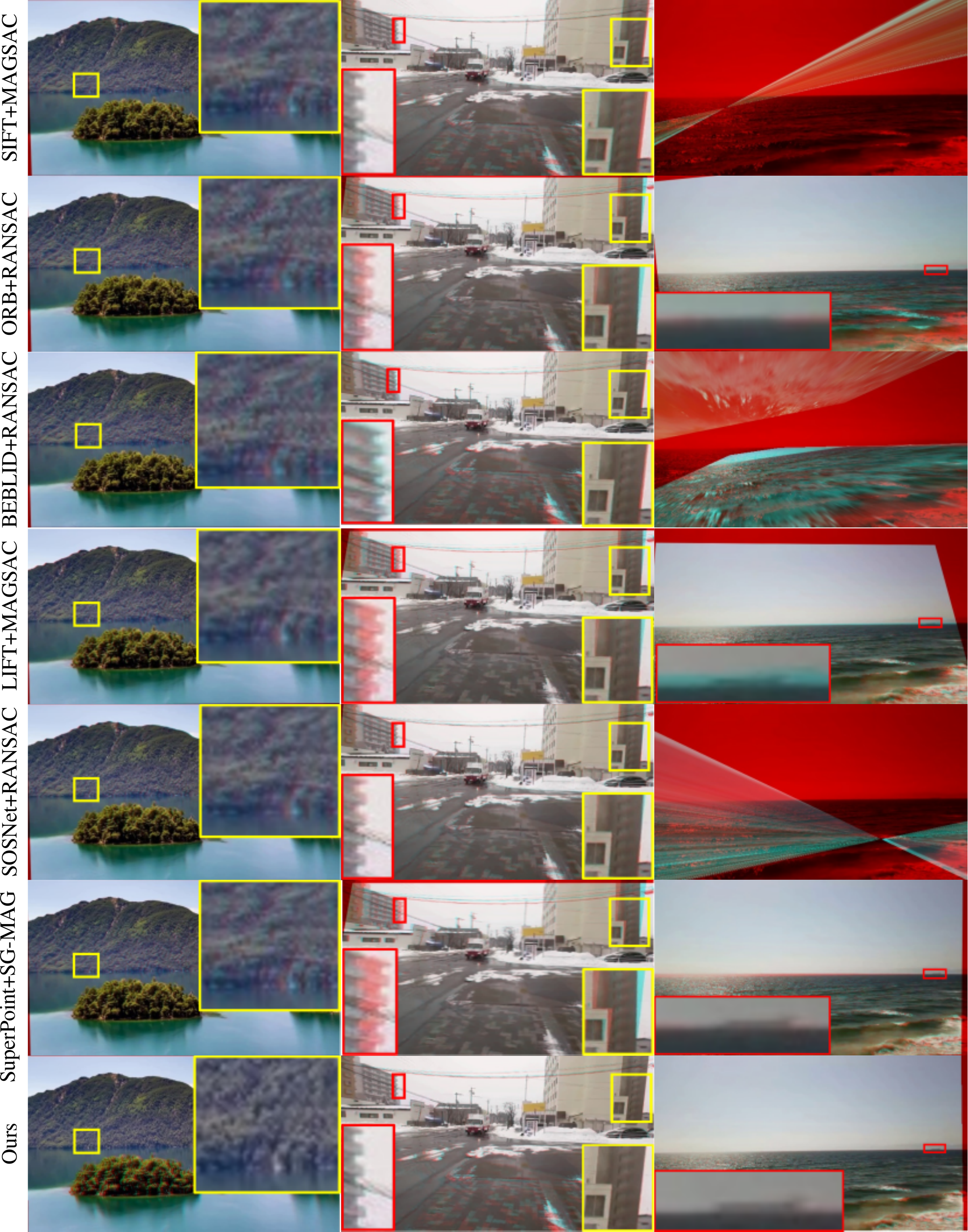}
    \vspace{-6mm}
	\caption{Qualitative results of feature-based methods and our method. For each feature-based method, we show its results with the best performed outlier rejection algorithm.}
	\vspace{-4mm}
	\label{fig:Featurebased}
\end{figure}

\begin{table*}[t]
	\centering
	\small
	\resizebox{0.98\linewidth}{!}{
		\begin{tabular}{
				r
				>{\arraybackslash}p{4.4cm}
				>{\arraybackslash}p{2.2cm}
				>{\arraybackslash}p{2.2cm}
				>{\arraybackslash}p{2.2cm}
				>{\arraybackslash}p{2.2cm}
				>{\arraybackslash}p{2.2cm}
				>{\arraybackslash}p{2.2cm}}
			\toprule
			1) &       & RE    & LT     & LL    & SF    & LF    & Avg \\
			\midrule
			2) & $\mathcal{I}_{3\times3}$                                                   &7.75(+2572.41\%)	&7.65(+1316.67\%)	&7.21(+1009.23\%)	&7.53(+1134.43\%)	&3.39(+726.83\%)	&6.70(+1240.00\%) \\ \midrule
			3) & SIFT~\cite{LoweSIFT04} + RANSAC~\cite{FischlerRansac81}                    &0.30(+3.45\%)	    &1.34(+148.15\%)	&4.03(+520.00\%)	&0.81(+32.79\%)	&0.57(+39.02\%)	&1.41(+182.00\%) \\ 
			4) & SIFT~\cite{LoweSIFT04} + MAGSAC~\cite{BarathMAGSAC19}                      &0.31(+6.90\%)	    &1.72(+218.52\%)	&3.39(+421.54\%)	&0.80(+31.15\%)	&0.47(+14.63\%)	&1.34(+168.00\%) \\  
			5) & ORB~\cite{RubleeRKB11ORB} + RANSAC~\cite{FischlerRansac81}                    &0.85(+193.10\%)	&2.59(+379.63\%)	&1.67(+156.92\%)	&1.10(+80.33\%)	&1.24(+202.44\%)	&1.48(+196.00\%)\\
			6) & ORB~\cite{RubleeRKB11ORB} + MAGSAC~\cite{BarathMAGSAC19}                     &0.97(+234.48\%)	&3.34(+518.52\%)	&1.58(+143.08\%)	&1.15(+88.52\%)	&1.4(+241.46\%)	&1.69(+238.00\%)\\
			7) & BEBLID~\cite{suarez2020beblid} + RANSAC~\cite{FischlerRansac81}           &0.78(+168.97\%)	&2.83(+424.07\%)	&1.38(+112.31\%)	&1.04(+70.49\%)	&1.33(+224.39\%)	&1.47(+194.00\%)\\
			8) & BEBLID~\cite{suarez2020beblid} + MAGSAC~\cite{BarathMAGSAC19}             &0.94(+224.14\%)	&3.73(+590.74\%)	&3.49(+436.92\%)	&1.17(+91.80\%)	&1.25(+204.88\%)	&2.12(+324.00\%) \\ \midrule
			9) & LIFT~\cite{YiTLF16} + RANSAC~\cite{FischlerRansac81}                      &0.40(+37.93\%)	&2.01(+272.22\%)	&1.14(+75.38\%)	&0.77(+26.23\%)	&0.68(+65.85\%)	&1.00(+100.00\%)\\
			10) & LIFT~\cite{YiTLF16} + MAGSAC~\cite{BarathMAGSAC19}                        &0.35(+20.69\%)	&1.85(+242.59\%)	&0.96(+47.69\%)	&0.72(+18.03\%)	&0.50(+21.95\%)	&0.88(+76.00\%)\\
			11) & SOSNet~\cite{TianSOSNet19} + RANSAC~\cite{FischlerRansac81}               &\textcolor{blue}{0.29(+0.00\%)}	&2.42(+348.15\%)	&3.71(+470.77\%)	&0.77(+26.23\%)	&0.59(+43.90\%)	&1.56(+212.00\%)\\
			12) & SOSNet~\cite{TianSOSNet19} + MAGSAC~\cite{BarathMAGSAC19}                 &0.30(+3.45\%)	&3.00(+455.56\%)	&3.66(+463.08\%)	&0.87(+42.62\%)	&0.49(+19.51\%)	&1.67(+234.00\%)\\
			13) & SuperPoint~\cite{detone2018superpoint} + RANSAC~\cite{FischlerRansac81}   &0.43(+48.28\%)	&0.85(+57.41\%)	&0.77(+18.46\%)	&0.84(+37.70\%)	&0.8(+95.12\%)	&0.74(+48.00\%)\\
			14) & SuperPoint~\cite{detone2018superpoint} + MAGSAC~\cite{BarathMAGSAC19}   	&0.45(+55.17\%)	&0.90(+66.67\%)	&0.77(+18.46\%)	&0.76(+24.59\%)	&0.67(+63.41\%)	&0.71(+42.00\%)\\
			15) & SuperPoint~\cite{detone2018superpoint}+SG-RAN~\cite{sarlin2020superglue}~\cite{FischlerRansac81} &0.41(+41.38\%)	&0.87(+61.11\%)	&0.72(+10.77\%)	&0.80(+31.15\%)	&0.75(+82.93\%)	&0.71(+42.00\%)\\
			16) & SuperPoint~\cite{detone2018superpoint} + SG-MAG~\cite{sarlin2020superglue}~\cite{BarathMAGSAC19}	&0.36(+24.14\%)	&0.79(+46.30\%)	&0.70(+7.69\%)	&0.71(+16.39\%)	&0.70(+70.73\%)	&0.63(+26.00\%) \\ \midrule
			17) & Supervised~\cite{detone2016deep} 	                                        &1.51(+420.69\%)	&4.48(+729.63\%)	&2.76(+324.62\%)	&2.62(+329.51\%)	&3.00(+631.71\%)	&2.87(+474.00\%) \\
			18) & Unsupervised~\cite{NguyenCSTK18}                                           &0.79(+172.41\%)	&2.45(+353.70\%)	&1.48(+127.69\%)	&1.11(+81.97\%)	&1.10(+168.29\%)	&1.39(+178.00\%) \\
			19) & CA-Unsupervised~\cite{zhang2020content}                                    &0.73(+151.72\%)	&1.01(+87.04\%)	&1.03(+58.46\%)	&0.92(+50.82\%)	&0.70(+70.73\%)	&0.88(+76.00\%) \\
			20) & BasesHomo~\cite{ye2021motion}                                              &\textcolor{blue}{0.29(+0.00\%)}	&\textcolor{blue}{0.54(+0.00\%)}	&\textcolor{blue}{0.65(+0.00\%)}	&\textcolor{blue}{0.61(+0.00\%)}	&\textcolor{blue}{0.41(+0.00\%)}	&\textcolor{blue}{0.50(+0.00\%)} \\ \midrule
			21) & HomoGAN (Ours)  &\textcolor{red}{0.22(-24.14\%)}	&\textcolor{red}{0.41(-24.07\%)}	&\textcolor{red}{0.57(-12.31\%)}	&\textcolor{red}{0.44(-27.87\%)}	&\textcolor{red}{0.31(-24.39\%)}	&\textcolor{red}{0.39(-22.00\%)} \\ 
			\bottomrule
		\end{tabular}
	}
    \vspace{-2mm}
	\caption{The point matching errors of our method and all comparison methods.  \textcolor{red}{Red} indicates the best result and \textcolor{blue}{blue} indicates the second best result. The percentages in the parentheses indicate the relative change in comparison to the second best result.
	}
	
	\label{tab:compare}
 	\vspace{-4mm}
\end{table*}%

\paragraph{Dataset}
Following \cite{ye2021motion} and \cite{zhang2020content}, we evaluate our method on a natural image dataset~\cite{zhang2020content} with 75.8k training pairs and 4.2k testing pairs of image size $320 \times 640$.
In both subsets, the image pairs are roughly evenly categorized into five types of scenes, respectively are regular (RE), low texture (LT), low light (LL), small foreground (SF), and large foreground (LF), where the last four are challenging scenes for homography estimation.
For evaluation, 6 pairs of ground-truth matching points are provided on each testing image.
We employ the average L2 distance from the predicted points to the ground-truth points on the target image as the evaluation metric.

\vspace{-4mm}
\paragraph{Implementation Details}
In training, we randomly crop patches of size $384 \times 512$ near the center of the original images as input to avoid out-of-bound coordinates after warping.
The number of scale levels is set to $k=3$.
Our network is implemented with PyTorch, and the training is performed on four NVIDIA RTX 2080Ti GPUs.
We employ the Adam optimizer~\cite{KingmaB14} with an initial learning rate of $1\times10^{-4}$ for model optimization, and it decays by a factor of 0.8 every epoch.
The batch size is 8.
The two stages of training take 10 and 2 epochs, respectively.
We reinitialize the learning rate to $1\times10^{-5}$ in the second stage.


\subsection{Comparison with Existing Methods}

\paragraph{Comparison methods}
We compare with three categories of existing homography estimation methods: 1) Traditional feature-based methods including SIFT~\cite{LoweSIFT04}, ORB~\cite{RubleeRKB11ORB} and BEBLID~\cite{suarez2020beblid}; 2) Learned feature-based methods including LIFT~\cite{YiTLF16}, SOSNet~\cite{TianSOSNet19} and SuperPoint~\cite{detone2018superpoint}; 3) Deep learning-based methods including Supervised~\cite{detone2016deep}, Unsupervised~\cite{NguyenCSTK18}, CA-Unsupervised~\cite{zhang2020content} and BasesHomo~\cite{ye2021motion}.
For all traditional and learned feature-based methods, we test them with two different outlier rejection algorithms RANSAC~\cite{FischlerRansac81} and MAGSAC~\cite{BarathMAGSAC19}, respectively.
Besides, SuperPoint is also tested with two customized rejection algorithms SuperGlue-RANSAC (SG-RAN) and SuperGlue-MAGSAC (SG-MAG)~\cite{sarlin2020superglue}.


\vspace{-4mm}
\paragraph{Qualitative comparison}
We first compare the qualitative results of HomoGAN with other methods.
In Fig.~\ref{fig:DNNbased}, we visualize the results of our method and four most related comparison methods, namely the deep learning-based methods, on three images with challenging scenes.
Fig.~\ref{fig:DNNbased}(a) is challenging because the plane of interest occupies a relatively small portion of the image, and it contains moving and still vehicles.
In Fig.~\ref{fig:DNNbased}(b), the large fountain results in significant depth disparity from the foreground to the background.
And Fig.~\ref{fig:DNNbased}(c) is a scene with low light and buildings in the distance.
As highlighted in the red and yellow boxes, existing methods cannot align these images as well as ours.
The Supervised~\cite{detone2016deep} method fails because it is trained on synthetic pairs without real depth disparity and dynamic contents, while the Unsupervised~\cite{detone2016deep} method predicts homographies based on the entire image, thus leading to inferior accuracy on the dominant plane.
CA-Unsupervised~\cite{zhang2020content}, and BasesHomo~\cite{ye2021motion} implicitly suppress undesired regions in their methods, but their performance is still limited by the lack of explicit guidance.
In contrast, our method could automatically focus on the dominant plane.


In Fig.~\ref{fig:Featurebased}, we also compare with feature-based methods. 
These feature methods are supposed to be robust to the plane-induced parallax with the help of outlier rejection algorithms. 
However, they still struggle in scenarios with blurry boundaries or low texture, such as the mountain and the sea in the 1st and 3rd columns of Fig.~\ref{fig:Featurebased}.
Without relying on keypoints, our method remains robust in these scenes.


\vspace{-4mm}
\paragraph{Quantitative comparison}
We report the quantitative results of all comparison methods in Table~\ref{tab:compare}, where rows 3-8 are traditional feature-based methods, rows 9-16 are learned feature-based methods, and rows 17-20 are deep learning-based methods. 
$\mathcal{I}_{3\times3}$ in the 1st row refers to the identity transformation, of which the error reflects the original distance between point pairs.

From Table~\ref{tab:compare}, we can see that our method achieves the state-of-the-art performance on all categories of the dataset and outperforms the best existing method BasesHomo by 22\%, with the matching error reduced from 0.50 to 0.39.
In regular (RE) scenes, feature-based methods usually perform well as these images are with high signal-noise ratio and provide sufficient features.
But our model still reduces the error on this category by $24.14\%$ compared to SOSNet+RANSAC. 
In low light (LL) and low texture (LT) scenes, most traditional feature-based methods fail to extract or match sufficient keypoints, resulting in unsatisfactory performance, while our method still has the lowest error among all.
It indicates the strong feature extraction ability of the proposed multi-scale transformer.

The small foreground (SF) and large foreground (LF) scenes are often accompanied by dynamic contents and multiple planes that cause problems for homography estimation. 
Compared with other deep learning-based methods with outlier rejection mechanisms, \textit{i.e.}, the implicitly generated mask in CA-Unsupervised~\cite{zhang2020content} and the low-rank representation in BasesHomo~\cite{ye2021motion}, our method significantly outperforms them in the LF and SF with errors reduced by at least $24.39\%$ and $27.87\%$, respectively.
It shows the superiority of our coplanarity-aware GAN in outlier rejection.

\begin{figure}[t]
	\centering
	\includegraphics[width=1\linewidth, height =0.45\linewidth]{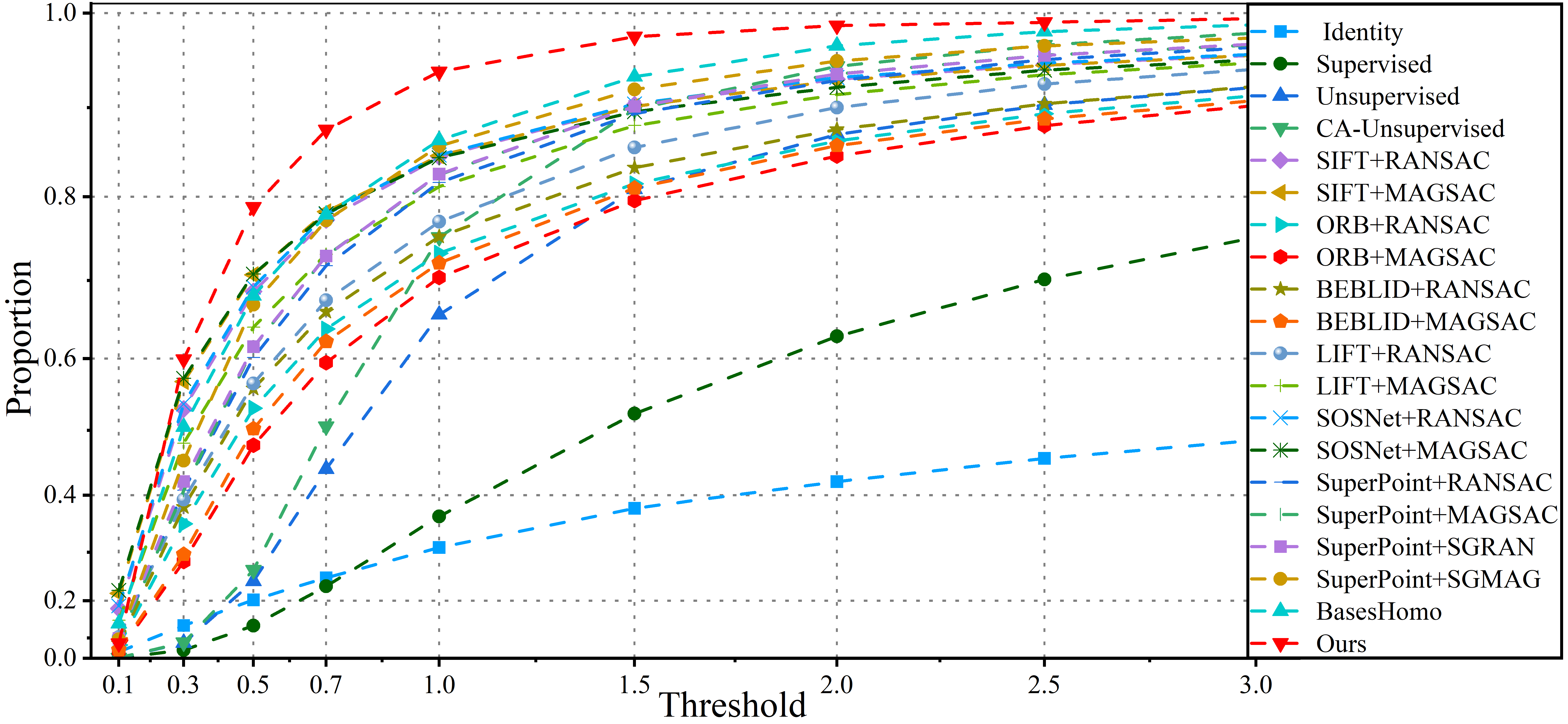}
    \vspace{-6mm}
	\caption{The proportion of inliers of all methods under various thresholds. Inliers indicate points with errors under the threshold.}
    \vspace{-2mm}
	\label{fig:robust}
\end{figure}

\vspace{-4mm}
\paragraph{Robustness evaluation}
To further investigate the robustness of all methods, we compute the proportion of inlier predictions with respect to a distance threshold. 
Specifically, for each method, we plot a curve where the axis $X$ is the distance threshold, and the axis $Y$ is the proportion of points with predictions errors less than the threshold, which are referred to as inliers. 
This curve could reflect the robustness of a method on homography estimation.
As shown in Fig.~\ref{fig:robust}, our method significantly surpasses other methods at most thresholds.
With a threshold of 1, our inlier proportion is $7.5\%$ higher than the second best ($93.9\%$ vs. $86.4\%$).

\begin{table*}[t]
	\centering
	\small
	\resizebox{0.98\linewidth}{!}{
		\begin{tabular}{
				r
				>{\arraybackslash}p{5cm}
				>{\arraybackslash}p{2cm}
				>{\arraybackslash}p{2cm}
				>{\arraybackslash}p{2cm}
				>{\arraybackslash}p{2cm}
				>{\arraybackslash}p{2cm}
				>{\arraybackslash}p{2cm}} 
			\toprule
			1) &Modification  & RE    & LT     & LL    & SF    & LF    & Avg \\
			\midrule
                    2) & Change to BasesHomo backbone  &0.29(+31.82$\%$)	&0.50(+21.95$\%$)	&0.63(+10.53$\%$)	&0.54(+22.73$\%$)	&0.36(+16.13$\%$)	&0.46(+17.95$\%$) \\
        3) & w/o weight token &0.23(+4.55$\%$)	&0.47(+14.63$\%$)	&0.66(+15.79$\%$)	&0.56(+27.27$\%$)	&0.37(+19.35$\%$)	&0.46(+17.95$\%$)  \\
        4) & w/o multi-scale &0.43(+95.45$\%$)	&1.01(+146.34$\%$)	&1.25(+119.30$\%$)	&1.13(+156.82$\%$)	&0.61(+96.77$\%$)	&0.89(+128.21$\%$) \\
        5) & w/o plane mask  &0.26(+18.18$\%$)	&0.59(+43.90$\%$)	&0.59(+3.51$\%$)	&0.63(+43.18$\%$)	&0.40(+29.03$\%$)	&0.49(+25.64$\%$)  \\ 
        6) & w/o coplanarity constraint   &0.24(+9.09$\%$)	&0.50(+21.95$\%$)	&0.64(+12.28$\%$)	&0.59(+34.09$\%$)	&0.36(+16.13$\%$)	&0.44(+12.82$\%$)  \\
        7) & Change to CA mask  &0.25(+13.64$\%$)	&0.66(+60.98$\%$)	&0.57(+0.00$\%$)	&0.54(+22.73$\%$)	&0.38(+22.58$\%$)	&0.48(+23.08$\%$)  \\ \midrule
        8) & Ours  &\textcolor{red}{0.22(+0.00$\%$)}	&\textcolor{red}{0.41(+0.00$\%$)}	&\textcolor{red}{0.57(+0.00$\%$)}	&\textcolor{red}{0.44(+0.00$\%$)}	&\textcolor{red}{0.31(+0.00$\%$)}	&\textcolor{red}{0.39(+0.00$\%$)} \\
			\bottomrule
		\end{tabular}
	}
	\vspace{-2mm}
	\caption{Results of ablation studies. Each row is the result of our method with a specific modification. Please refer to the text for details.
	}
	\vspace{-4mm}
	\label{tab:ablation}
\end{table*}

\subsection{Ablation Studies}


\paragraph{Homography Estimation Transformer}
To demonstrate the ability of the proposed transformer network in homography estimation, we change it to the backbone of BasesHomo~\cite{ye2021motion}, which is a ResNet-34 architecture with customized Low Rank Representation blocks, and achieves the second best result in Table~\ref{tab:compare}.
By comparing row $2$ with row $8$ in Table~\ref{tab:ablation}, we can see that the average error of our method increases from $0.39$ to $0.46$ with this change.
This result demonstrates the superiority of the proposed transformer over CNNs in homography estimation.
Meanwhile, the numbers of parameters in our transformer ($2.045$M) is much lower than it of the BasesHomo backbone ($21.296$M).

From another perspective, this experiment also demonstrates that our coplanarity-aware GAN is applicable to different homography estimators, because the BasesHomo backbone with our GAN achieves an average error of $0.46$, which is lower than its original error of $0.50$.

\vspace{-4mm}
\paragraph{Weight token}
In the class-attention decoder of our transformer, we employ a weight token to summarize weight-aware information from the self-attention feature for homography estimation.
In this experiment, we remove this token from our network and directly fed the self-attention feature into an MLP to predict the homography. The results are reported in row $3$ of Table~\ref{tab:ablation}.
Comparing row $3$ with row $8$, we can see that the error increases by $17.95\%$ from $0.39$ to $0.46$.
It indicates that using an independent learnable token to summarize global information in the decoding stage is beneficial for homography estimation.


\begin{figure}[t]
	\centering
	\includegraphics[width=0.95\linewidth]{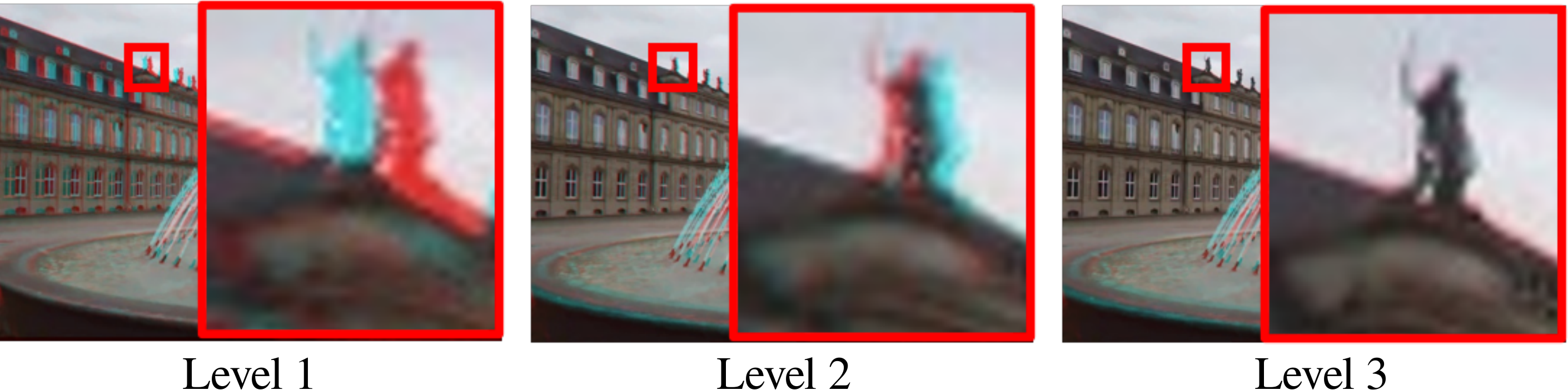}
    \vspace{-2mm}
	\caption{Results of the multi-scale transformer at each level. It shows how the homography is predicted from coarse to fine.}
	\label{fig:multi-scale}
	\vspace{-4mm}
\end{figure}

\vspace{-4mm}
\paragraph{Multi-scale architecture}
In the transformer network, we use three consecutive transformer modules to predict the homography from coarse to fine.
In this experiment, we change to using only one module to directly predict the final homography to validate the effectiveness of the multi-scale architecture.
From row 4 of Table~\ref{tab:ablation}, we find that when using only one transformer module, the average error increases to 0.89, which is significantly higher than the error of 0.39 when using three modules.
This result shows that bridging the rich-semantic features at the high level with high-resolution features at the low level in a coarse-to-fine fashion is beneficial for homography estimation.
Besides, we visualize the alignment results after each transformer module in Fig.~\ref{fig:multi-scale}.
It illustrates how the predicted homographies at different levels progressively align two images.


\vspace{-4mm}
\paragraph{Plane mask}
To validate the usefulness of the generated plane mask, we remove all mask related operations from our network and check the performance, which is exactly the result of the first stage of training.
With only the first stage training, the average error of our network is 0.49, as reported in row $5$ of Table~\ref{tab:ablation}, which is already better than the previous SOTA, but can still be reduced.
After adding the mask related operations back to the network and fine-tuning for 2 more epochs, we further reduce the average error to 0.39.
It clearly shows the usefulness of our plane mask.


\begin{figure}[t]
	\centering
	\includegraphics[width=0.95\linewidth]{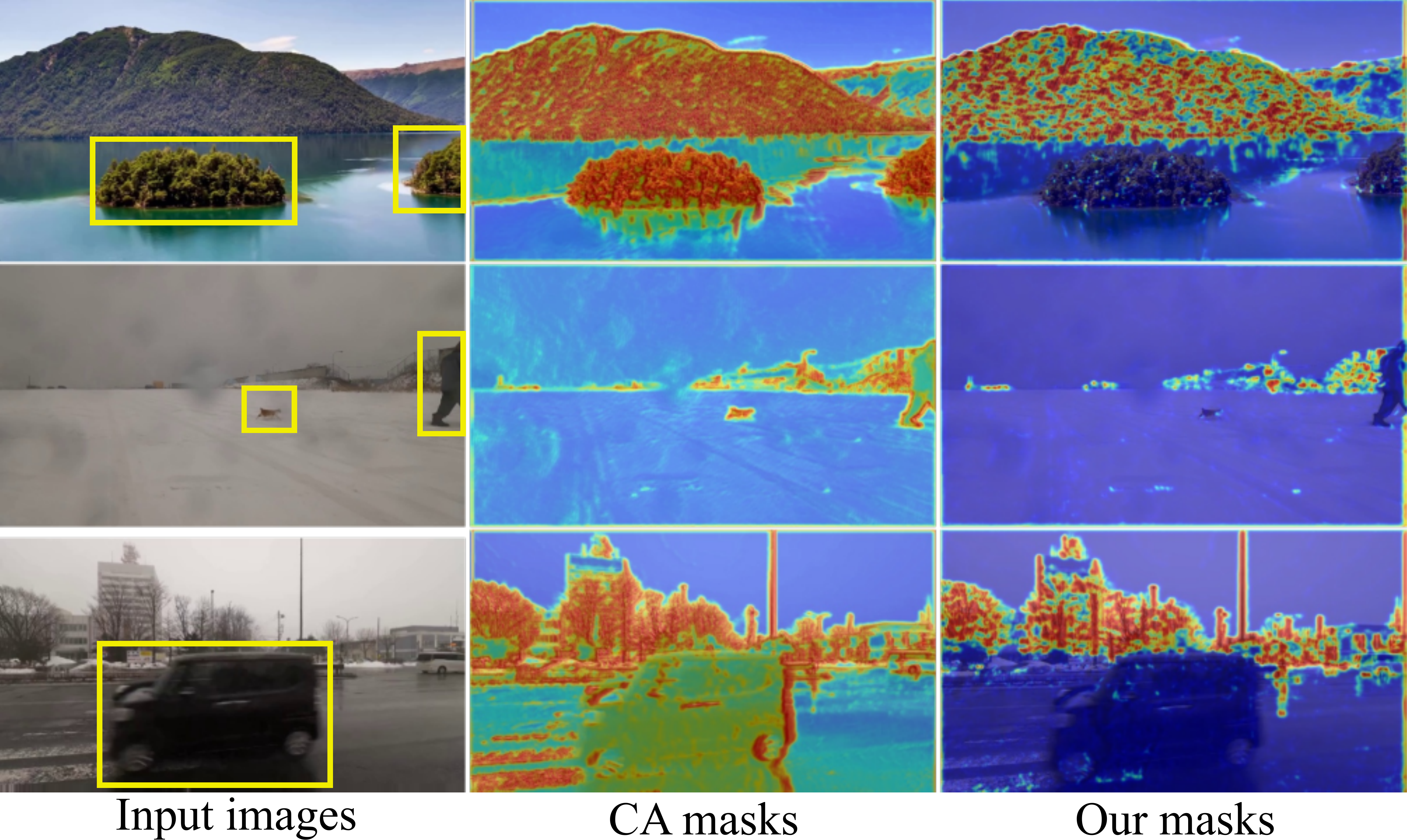}
    \vspace{-2mm}
	\caption{Masks predicted by CA-Unsupervised~\cite{zhang2020content} and our method. With the coplanarity constraint, our masks are able to focus on a dominant plane. Best viewed in color.}
	\vspace{-5mm}
	\label{fig:mask_compare}
\end{figure}

\vspace{-4mm}
\paragraph{Coplanarity constraint}
Mask prediction for unsupervised homography estimation has been introduced by~\cite{zhang2020content}.
But in this work, we propose to impose the coplanarity constraint to the mask to make it focus on the dominant plane, which is realized by the coplanarity-aware GAN.
In this experiment, we try to use different methods to generate the mask to validate the necessity of coplanarity constraint.
First, we remove the discriminator and the adversarial loss from our network training, such that the mask is generated without coplanarity constraint.
Second, we change the mask generation method to be the same as~\cite{zhang2020content}, in which the masks are generated from the output of the feature projector and then applied to the extracted features and the triplet loss. It also lacks the coplanarity constraint.
The results of our method with these two mask generation approaches are reported in row 6 and row 7 of Table~\ref{tab:ablation}.
Comparing rows 6, 7, and 8, we can see that the mask generated by our coplanarity-aware GAN achieves the best performance among the three. 
It indicates that imposing an explicit coplanarity constraint is more effective than implicit mask generation in homography estimation.
Moreover, we display the masks generated by ~\cite{zhang2020content} and our method on three representative images in Fig.~\ref{fig:mask_compare}.
The visualizations show that our method could generate masks that focus on the dominant plane without interferences from foreground objects in various scenes.



\section{Conclusion}
\label{sec:conclusion}
We have presented HomoGAN for unsupervised homography estimation. We noticed the problem of plane-induced parallax when learning homography without constraints, and proposed a coplanarity-aware GAN to solve it. Compared to previous methods, our method could generate a dominant plane mask with explicit coplanarity constraint, thus guiding the homography estimator to focus on the dominant plane. Besides, a multi-scale transformer network has been proposed to estimate the homography from coarse to fine, which has gained improvement over previous CNN-based estimators. With these two designs, we have achieved the SOTA performance on the standard benchmark.

\section*{Acknowledgement}
This work was supported in part by the National Natural Science Foundation of China (NSFC) under grants (No. 62176170, 61872067, 62066042 and 62172032) and in part by Sichuan Province Key Research and Development Project (No.2020YJ0282).

{\small
\bibliographystyle{ieee_fullname}
\bibliography{homo}

\begin{thebibliography}{10}\itemsep=-1pt

\bibitem{BarathMAGSAC19}
Daniel Barath, Jiri Matas, and Jana Noskova.
\newblock {MAGSAC:} marginalizing sample consensus.
\newblock In {\em {Proc. CVPR}}, pages 10197--10205, 2019.

\bibitem{BayETGSURF08}
Herbert Bay, Andreas Ess, Tinne Tuytelaars, and Luc~Van Gool.
\newblock Speeded-up robust features {(SURF)}.
\newblock {\em Comput. Vis. Image Underst.}, 110(3):346--359, 2008.

\bibitem{BayTGSURF06}
Herbert Bay, Tinne Tuytelaars, and Luc~Van Gool.
\newblock {SURF:} speeded up robust features.
\newblock In {\em {Proc. ECCV}}, pages 404--417, 2006.

\bibitem{bian2017gms}
JiaWang Bian, Wen-Yan Lin, Yasuyuki Matsushita, Sai-Kit Yeung, Tan-Dat Nguyen,
  and Ming-Ming Cheng.
\newblock Gms: Grid-based motion statistics for fast, ultra-robust feature
  correspondence.
\newblock In {\em {Proc. CVPR}}, pages 4181--4190, 2017.

\bibitem{chen2017deeplab}
Liang-Chieh Chen, George Papandreou, Iasonas Kokkinos, Kevin Murphy, and Alan~L
  Yuille.
\newblock Deeplab: Semantic image segmentation with deep convolutional nets,
  atrous convolution, and fully connected crfs.
\newblock {\em IEEE transactions on pattern analysis and machine intelligence},
  40(4):834--848, 2017.

\bibitem{conrad2010homography}
Daniel Conrad and Guilherme~N DeSouza.
\newblock Homography-based ground plane detection for mobile robot navigation
  using a modified em algorithm.
\newblock In {\em 2010 IEEE International Conference on Robotics and
  Automation}, pages 910--915. IEEE, 2010.

\bibitem{de2016dominant}
JA de Jes{\'u}s Osuna-Coutino, Jose Martinez-Carranza, Miguel Arias-Estrada,
  and Walterio Mayol-Cuevas.
\newblock Dominant plane recognition in interior scenes from a single image.
\newblock In {\em 2016 23rd International Conference on Pattern Recognition
  (ICPR)}, pages 1923--1928. IEEE, 2016.

\bibitem{detone2016deep}
Daniel DeTone, Tomasz Malisiewicz, and Andrew Rabinovich.
\newblock Deep image homography estimation.
\newblock {\em arXiv preprint arXiv:1606.03798}, 2016.

\bibitem{detone2018superpoint}
Daniel DeTone, Tomasz Malisiewicz, and Andrew Rabinovich.
\newblock Superpoint: Self-supervised interest point detection and description.
\newblock In {\em {Proc. CVPRW}}, pages 224--236, 2018.

\bibitem{FischlerRansac81}
Martin~A. Fischler and Robert~C. Bolles.
\newblock Random sample consensus: {A} paradigm for model fitting with
  applications to image analysis and automated cartography.
\newblock {\em Commun. {ACM}}, 24(6):381--395, 1981.

\bibitem{ganin2015GRL}
Yaroslav Ganin and Victor Lempitsky.
\newblock Unsupervised domain adaptation by backpropagation.
\newblock In {\em International conference on machine learning}, pages
  1180--1189. PMLR, 2015.

\bibitem{gelfand2010multi}
Natasha Gelfand, Andrew Adams, Sung~Hee Park, and Kari Pulli.
\newblock Multi-exposure imaging on mobile devices.
\newblock In {\em Proceedings of the 18th ACM international conference on
  Multimedia}, pages 823--826, 2010.

\bibitem{gulrajani2017WGANGP}
Ishaan Gulrajani, Faruk Ahmed, Martin Arjovsky, Vincent Dumoulin, and Aaron
  Courville.
\newblock Improved training of wasserstein gans.
\newblock {\em arXiv preprint arXiv:1704.00028}, 2017.

\bibitem{guo2016joint}
Heng Guo, Shuaicheng Liu, Tong He, Shuyuan Zhu, Bing Zeng, and Moncef Gabbouj.
\newblock Joint video stitching and stabilization from moving cameras.
\newblock {\em IEEE Transactions on Image Processing}, 25(11):5491--5503, 2016.

\bibitem{daglib_AHAZ}
Andrew Harltey and Andrew Zisserman.
\newblock {\em Multiple view geometry in computer vision {(2.} ed.)}.
\newblock 2006.

\bibitem{1977RobustIRLS}
Paul~W. Holland and Roy~E. Welsch.
\newblock Robust regression using iteratively reweighted least-squares.
\newblock {\em Communications in Statistics}, 6(9):813--827, 1977.

\bibitem{KingmaB14}
Diederik~P. Kingma and Jimmy Ba.
\newblock Adam: {A} method for stochastic optimization.
\newblock In {\em {Proc. ICLR}}, 2015.

\bibitem{LeLZA20}
Hoang Le, Feng Liu, Shu Zhang, and Aseem Agarwala.
\newblock Deep homography estimation for dynamic scenes.
\newblock In {\em {Proc. CVPR}}, pages 7649--7658, 2020.

\bibitem{liu2019planercnn}
Chen Liu, Kihwan Kim, Jinwei Gu, Yasutaka Furukawa, and Jan Kautz.
\newblock Planercnn: 3d plane detection and reconstruction from a single image.
\newblock In {\em Proceedings of the IEEE/CVF Conference on Computer Vision and
  Pattern Recognition}, pages 4450--4459, 2019.

\bibitem{liu2018planenet}
Chen Liu, Jimei Yang, Duygu Ceylan, Ersin Yumer, and Yasutaka Furukawa.
\newblock Planenet: Piece-wise planar reconstruction from a single rgb image.
\newblock In {\em Proceedings of the IEEE Conference on Computer Vision and
  Pattern Recognition}, pages 2579--2588, 2018.

\bibitem{liu2021swin}
Ze Liu, Yutong Lin, Yue Cao, Han Hu, Yixuan Wei, Zheng Zhang, Stephen Lin, and
  Baining Guo.
\newblock Swin transformer: Hierarchical vision transformer using shifted
  windows.
\newblock {\em arXiv preprint arXiv:2103.14030}, 2021.

\bibitem{LoweSIFT04}
David~G. Lowe.
\newblock Distinctive image features from scale-invariant keypoints.
\newblock {\em Int. J. Comput. Vis.}, 60(2):91--110, 2004.

\bibitem{ma2019localityLPM}
Jiayi Ma, Ji Zhao, Junjun Jiang, Huabing Zhou, and Xiaojie Guo.
\newblock Locality preserving matching.
\newblock {\em {International Journal of Computer Vision}}, 127(5):512--531,
  2019.

\bibitem{Mur-ArtalMT15}
Raul Mur{-}Artal, J.~M.~M. Montiel, and Juan~D. Tard{\'{o}}s.
\newblock {ORB-SLAM:} {A} versatile and accurate monocular {SLAM} system.
\newblock {\em {IEEE} Trans. Robotics}, 31(5):1147--1163, 2015.

\bibitem{NguyenCSTK18}
Ty Nguyen, Steven~W. Chen, Shreyas~S. Shivakumar, Camillo~Jose Taylor, and
  Vijay Kumar.
\newblock Unsupervised deep homography: {A} fast and robust homography
  estimation model.
\newblock {\em {IEEE} Robotics Autom. Lett.}, 3(3):2346--2353, 2018.

\bibitem{RubleeRKB11ORB}
Ethan Rublee, Vincent Rabaud, Kurt Konolige, and Gary~R. Bradski.
\newblock {ORB:} an efficient alternative to {SIFT} or {SURF}.
\newblock In {\em {Proc. ICCV}}, pages 2564--2571, 2011.

\bibitem{sarlin2020superglue}
Paul-Edouard Sarlin, Daniel DeTone, Tomasz Malisiewicz, and Andrew Rabinovich.
\newblock Superglue: Learning feature matching with graph neural networks.
\newblock In {\em {Proc. CVPR}}, pages 4938--4947, 2020.

\bibitem{shao2021localtrans}
Ruizhi Shao, Gaochang Wu, Yuemei Zhou, Ying Fu, Lu Fang, and Yebin Liu.
\newblock Localtrans: A multiscale local transformer network for
  cross-resolution homography estimation.
\newblock {\em arXiv preprint arXiv:2106.04067}, 2021.

\bibitem{suarez2020beblid}
Iago Su{\'a}rez, Ghesn Sfeir, Jos{\'e}~M Buenaposada, and Luis Baumela.
\newblock Beblid: Boosted efficient binary local image descriptor.
\newblock {\em Pattern Recognition Letters}, 133:366--372, 2020.

\bibitem{sun2021loftr}
Jiaming Sun, Zehong Shen, Yuang Wang, Hujun Bao, and Xiaowei Zhou.
\newblock {LoFTR}: Detector-free local feature matching with transformers.
\newblock 2021.

\bibitem{tan2021planetr}
Bin Tan, Nan Xue, Song Bai, Tianfu Wu, and Gui-Song Xia.
\newblock Planetr: Structure-guided transformers for 3d plane recovery.
\newblock In {\em Proceedings of the IEEE/CVF International Conference on
  Computer Vision}, pages 4186--4195, 2021.

\bibitem{TianSOSNet19}
Yurun Tian, Xin Yu, Bin Fan, Fuchao Wu, Huub Heijnen, and Vassileios Balntas.
\newblock Sosnet: Second order similarity regularization for local descriptor
  learning.
\newblock In {\em {Proc. CVPR}}, pages 11016--11025, 2019.

\bibitem{touvron2021cait}
Hugo Touvron, Matthieu Cord, Alexandre Sablayrolles, Gabriel Synnaeve, and
  Herv{\'e} J{\'e}gou.
\newblock Going deeper with image transformers.
\newblock {\em arXiv preprint arXiv:2103.17239}, 2021.

\bibitem{yang2018recovering}
Fengting Yang and Zihan Zhou.
\newblock Recovering 3d planes from a single image via convolutional neural
  networks.
\newblock In {\em Proceedings of the European Conference on Computer Vision
  (ECCV)}, pages 85--100, 2018.

\bibitem{ye2021motion}
Nianjin Ye, Chuan Wang, Haoqiang Fan, and Shuaicheng Liu.
\newblock Motion basis learning for unsupervised deep homography estimation
  with subspace projection.
\newblock In {\em Proceedings of the IEEE/CVF International Conference on
  Computer Vision (ICCV)}, pages 13117--13125, October 2021.

\bibitem{YiTLF16}
Kwang~Moo Yi, Eduard Trulls, Vincent Lepetit, and Pascal Fua.
\newblock {LIFT:} learned invariant feature transform.
\newblock In {\em {Proc. ECCV}}, volume 9910, pages 467--483, 2016.

\bibitem{ZaragozaAPAP13}
Julio Zaragoza, Tat{-}Jun Chin, Michael~S. Brown, and David Suter.
\newblock As-projective-as-possible image stitching with moving {DLT}.
\newblock In {\em {Proc. CVPR}}, pages 2339--2346, 2013.

\bibitem{zhang2019learningOAN}
Jiahui Zhang, Dawei Sun, Zixin Luo, Anbang Yao, Lei Zhou, Tianwei Shen, Yurong
  Chen, Long Quan, and Hongen Liao.
\newblock Learning two-view correspondences and geometry using order-aware
  network.
\newblock In {\em {Proc. ICCV}}, pages 5845--5854, 2019.

\bibitem{zhang2020content}
Jirong Zhang, Chuan Wang, Shuaicheng Liu, Lanpeng Jia, Nianjin Ye, Jue Wang, Ji
  Zhou, and Jian Sun.
\newblock Content-aware unsupervised deep homography estimation.
\newblock In {\em {Proc. ECCV}}, pages 653--669, 2020.

\bibitem{zhang2000flexible}
Zhengyou Zhang.
\newblock A flexible new technique for camera calibration.
\newblock {\em IEEE Transactions on pattern analysis and machine intelligence},
  22(11):1330--1334, 2000.

\bibitem{zou2012coslam}
Danping Zou and Ping Tan.
\newblock Coslam: Collaborative visual slam in dynamic environments.
\newblock {\em IEEE transactions on pattern analysis and machine intelligence},
  35(2):354--366, 2012.

\end{thebibliography}
}

\end{document}